\documentclass[letterpaper]{article} 
\usepackage{aaai23}  
\usepackage{times}  
\usepackage{helvet}  
\usepackage{courier}  
\usepackage[hyphens]{url}  
\usepackage{graphicx} 
\usepackage{natbib}  
\usepackage{caption} 
\usepackage{algorithm}
\usepackage{algorithmic}

\usepackage{newfloat}
\usepackage{listings}

\usepackage{amsmath}
\usepackage{multirow}
\usepackage{amssymb}
\usepackage{booktabs}
\usepackage{subfigure}

\DeclareCaptionStyle{ruled}{labelfont=normalfont,labelsep=colon,strut=off} 
\floatstyle{ruled}
\newfloat{listing}{tb}{lst}{}
\floatname{listing}{Listing}
\usepackage{hyperref} 

\title{Compositional Prototypical Networks for Few-Shot Classification}
\author{
    Qiang Lyu,
    Weiqiang Wang\thanks{Corresponding author.}
}
\affiliations{

    School of Computer Science and Technology, University of Chinese Academy of Sciences\\

    luqiang20@mails.ucas.ac.cn, wqwang@ucas.ac.cn
}

\usepackage{bibentry}
\begin{document}

\maketitle
    \begin{abstract}
    It is assumed that pre-training provides the feature extractor with strong class transferability and that high novel class generalization can be achieved by simply reusing the transferable feature extractor. In this work, our motivation is to explicitly learn some fine-grained and transferable meta-knowledge so that feature reusability can be further improved. Concretely, inspired by the fact that humans can use learned concepts or components to help them recognize novel classes, we propose Compositional Prototypical Networks (CPN) to learn a transferable prototype for each human-annotated attribute, which we call a component prototype. We empirically demonstrate that the learned component prototypes have good class transferability and can be reused to construct compositional prototypes for novel classes. Then a learnable weight generator is utilized to adaptively fuse the compositional and visual prototypes. Extensive experiments demonstrate that our method can achieve state-of-the-art results on different datasets and settings. The performance gains are especially remarkable in the 5-way 1-shot setting. The code is available at https://github.com/fikry102/CPN.
    \end{abstract}
    
    \section{Introduction}
    Deep learning models have made tremendous progress in many computer vision tasks such as image classification~\cite{AlexNet,ResNet}, object detection~\cite{R-CNN,Mask_R-CNN}, and semantic segmentation~\cite{FCN}. However, such models heavily rely on large-scale labeled data, which can be impractical or laborious to collect.
    Humans, by contrast, can easily learn to recognize novel classes
    from only one or a few examples by utilizing prior knowledge and experience. 
    To bridge the gap between deep learning models and human intelligence, 
    few-shot learning~\cite{FSL_FF, FSL_Miller} (FSL) has recently received much attention from researchers. 
    Inspired by human learning behavior, FSL aims to recognize unlabeled examples from 
    novel classes using only a few labeled examples, with prior knowledge learned
    from the disjoint base classes dataset containing abundant examples per class.
    
    \begin{figure}[!ht]
        \centering
      \includegraphics[width=0.46\textwidth]{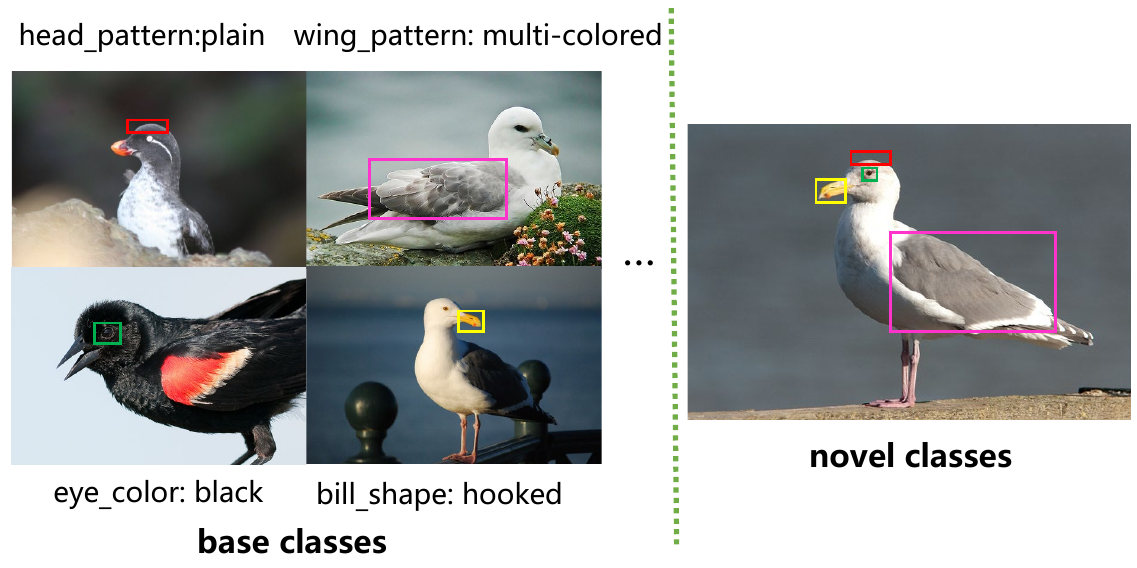}
        \caption{An illustration of the reusable components which can help humans recognize new birds better. We propose to learn component prototypes for human-annotated attributes from base classes and reuse them to construct compositional prototypes for novel classes.}
        \label{figure_reuse}
    \end{figure}
    
    Meta-learning~\cite{schmidhuber1987evolutionary}, or learning to learn, has become a
    popular learning paradigm for few-shot learning. Meta-learning based approaches aim to 
    extract transferable meta-knowledge from collections of learning tasks so that
    the learned model can quickly adapt to new tasks using the learned meta-knowledge. 
    A recent work~\cite{raghu_rapid} has demonstrated that the reusable features, i.e., 
    the high-quality features contained in the learned initial condition, are the dominant factor for the effectiveness of Model Agnostic Meta-Learning~\cite{MAML_Finn} (MAML),
    a well-known meta-learning method.  
    In addition to MAML, a succession of meta-learning based approaches 
    have been proposed to obtain various meta-knowledge, e.g.,
    a good initialization~\cite{meta_learning_LSTM,MAML_Finn},
    an optimization algorithm~\cite{GD_by_GD,meta_learning_LSTM},
    or a meta-structure~\cite{FEAT,Cross_attention_FSL}. 
    
    However, many recent works~\cite{Meta_baseline,A_closer_look,Rethinking_FSL,baseline_FSL} 
    indicate that pre-training a feature extractor on the whole base classes dataset can achieve comparable or even better performance 
    than many existing meta-learning methods. Some researches 
    ~\cite{global_FSL,Meta_baseline} show that global classification
    in the pre-training stage can provide the feature extractor with strong
    class transferability and therefore is theoretically 
    beneficial to the subsequent meta-learning stage.
    Here, we argue that although high novel class generalization can be 
    achieved by reusing the pre-trained feature extractor, there is still a huge
    gap compared to how humans reuse prior knowledge and experience.
    More specifically, humans can summarize some reusable concepts or 
    components from past learning tasks and then draw connections between 
    these concepts or components with new learning tasks 
    to help them learn new things better. For instance, if a child has ever seen several birds, one with multi-colored wings, one with black eyes, one with a hooked beak, and so on, 
    he can easily learn to recognize a new kind of bird with these features (see Figure~\ref{figure_reuse}).
    
    Based on the above analysis, our motivation is to explicitly learn some fine-grained and transferable meta-knowledge from base classes
    and then reuse the learned meta-knowledge to 
    recognize novel classes 
    with only a few labeled examples. 
    As some previous works~\cite{AGAM_attention_alignment,Comp_rep,MAP,AM3,ASL}, 
    our work also utilizes \textbf{category-level} attribute annotations, \emph{i.e.}, 
    only one attribute score vector for each class. 
    AM3~\cite{AM3} proposes a modality mixture mechanism 
    that can adaptively combine visual and semantic information. AGAM~\cite{AGAM_attention_alignment} proposes an attention alignment mechanism to align the self-guided branch's channel attention and spatial attention with the attributes-guided branch. 
    MAP-Net~\cite{MAP} designs a Modal-Alternating Propagation
    Module (MAP-Module) to alleviate the information asymmetry problem between semantic-guided and nonsemantic-guided examples. However, these methods
    simply take the attribute score vector as a whole and do not 
    pay enough attention to the more fine-grained attributes.  
    Although the method proposed in ~\cite{Comp_rep} has considered fine-grained attributes, 
    it aims to use a regularization technique to improve the compositionality of  
    learned representations. By contrast, our work aims to learn transferable component prototypes for
    human-annotated attributes and then use these learned component prototypes to construct compositional prototypes
    for novel classes.

    \section{Related Work}
    \paragraph{Few-Shot Learning.} 
    Few-shot learning has been widely studied in recent years.
    Existing few-shot learning methods can be roughly classified 
    into three branches: optimization-based methods, metric-based methods, and semantic-based methods. Our work belongs to both metric-based and
    semantic-based methods. 
    
    Optimization-based methods~\cite{MAML_Finn,meta_learning_LSTM,GD_by_GD} 
    aim to learn a good initialization or an optimization
    algorithm so that the model can quickly adapt to new tasks 
    with only a few steps of gradient descent. 
    MAML~\cite{MAML_Finn} combines a second-order optimizing strategy with the meta-learning framework. To overcome difficulties caused by direct optimization in high-dimensional parameter spaces, LEO~\cite{LEO} performs meta-learning in a low-dimensional latent space from which high-dimensional parameters can be generated.
    
    Metric-based methods aim to learn a good embedding function
    so that the embedded examples from novel classes can be correctly classified
    using a proper distance metric. For instance, 
    Prototypical Networks~\cite{Prototypical_Net} calculate the mean vector of the embedded examples of a given class as the prototype of this class and 
    choose the Euclidean distance as its distance metric. Furthermore, Relation Networks~\cite{Relation_Network} use a learnable distance metric, which can be jointly optimized with the embedding function. DeepEMD~\cite{DeepEMD} employs the Earth Mover's Distance~\cite{EMD} (EMD) as its distance metric and calculates the similarity between two images from the perspective of local features.
    
    Unlike the methods that rely solely on visual information, semantic-based methods resort to auxiliary semantic information. 
    For instance, AGAM~\cite{AGAM_attention_alignment} and MAP-Net~\cite{MAP} use category-level attribute score vectors. COMET~\cite{Concept_Learner} 
    uses the human-annotated location coordinates for predefined parts of birds to extract fine-grained features for these parts. By contrast, our method focuses on learning \textbf{reusable} component prototypes for predefined attributes. RS-FSL~\cite{afham2021rich} replaces numerical class labels with category-level language descriptions.
    Prototype Completion Network~\cite{Prototype_Completion} utilizes diverse 
    semantic information to learn to complete prototypes, namely, class parts/attributes extracted from WordNet and word embeddings calculated by GloVe~\cite{Glove} of all categories and attributes.
    By contrast, our work only utilizes category-level attribute score vectors as auxiliary semantic information following AGAM~\cite{AGAM_attention_alignment} and MAP-Net~\cite{MAP},.

    \paragraph{Zero-Shot Learning.}
    Zero-shot learning (ZSL) also aims to classify examples from unseen classes. Compared to few-shot learning, there is only semantic information (e.g., attribute annotations or word embeddings) and no labeled images for unseen classes in ZSL. Therefore the key insight of ZSL is to transfer semantic knowledge from seen classes to unseen classes. 
    
    Early ZSL methods~\cite{Discriminative_ZSL} learn an embedding function from visual space to semantic space. Then the unlabeled examples for unseen classes can be projected into semantic space 
    in which category-level attribute vectors reside. However, these methods train their models only
    using examples from seen classes, and the learned models are inevitably biased towards seen classes when it comes to the generalized ZSL setting. 
    To reduce the bias problem, some researches~\cite{verma2018generalized,xian2019f} utilize generative models to 
    generate images or visual features for unseen classes based on the category-level attribute vectors. Furthermore, some recent methods~\cite{chen2021hsva,schonfeld2019generalized} learn to map visual and semantic features into a joint space. 
    Our work is somewhat inspired by such methods since we adaptively fuse the semantic compositional prototype and the visual prototype.
    
    \paragraph{Compositional Representations.}
    It is considered that humans can harness compositionality to rapidly acquire and generalize knowledge to new tasks or situations in cognitive science literature~\cite{hoffman1984parts,biederman1987recognition,lake2017building}.
    Compositional representations allow learning new concepts from a few examples by composing learned primitives, which is a desired property for few-shot learning approaches.
    
    Andreas \emph{et al.}~\cite{andreas2018measuring} propose a method to evaluate 
    the compositionality of the learned representations by measuring how well they
    can be approximated by the composition of a group of primitives.
    Following~\cite{andreas2018measuring}, Tokmakov \emph{et al.}~\cite{Comp_rep} design a regularization technique to improve the compositionality of learned representations.
    ConstellationNet~\cite{Constellation_Net} uses self-attention mechanisms 
    to model the relation between cell features and utilize the K-means algorithm to conduct cell feature clustering. The learned cluster centers can be viewed as 
    potential object parts. Similarly, CPDE~\cite{primitive_discovery} learns primitives related to object parts by self-supervision and uses an Enlarging-Reducing loss (ER loss) to enlarge the activation of important primitives and reduce that of others.
    These inspiring works motivate us to integrate the idea of compositional representations into our few-shot learning method. Concretely, the compositional prototypes constructed by learned component prototypes in our work can be regarded as a kind of compositional representation.
    
    \section{Method}
    \subsection{Preliminaries}
    \subsubsection{Problem Formulation}
    In standard few-shot classification, there are two mutually exclusive class sets, 
    base classes set $C_{base}$ and novel classes set $C_{novel}$, where $C_{base}\cap C_{novel}=\emptyset$
    . In $N$-way $K$-shot 
    setting~\cite{Matching_Net}, we first sample $N$ categories from $C_{novel}$, and then the support set 
    $\mathcal{S} = \left\{\left(x_{i}, z_{i}, y_{i}\right)|y_i \in C_{novel} \right\}_{i=1}^{N \times K}$ 
    is constructed by sampling $K$ examples for each of the $N$ categories.
    Here, $x_i$ is the $i$-th image, $y_i$ denotes the 
    class label of the image, and $z_{i}$ denotes the attribute score 
    vector of the image, where each dimension of $z_{i}$ 
    corresponds to an attribute in a predefined attribute set $\mathcal{A}=\{a_j\}_{j=1}^M$. Similarly, the query set 
    $\mathcal{Q}=\left\{\left(x_{i}, y_{i}\right)|y_i \in C_{novel} \right\}_{i=1}^{N\times Q}$ 
    contains $Q$ examples for each of the $N$ categories.
    It is worth noting that the attribute score vector is unavailable for examples in the query set.
    An episode (task) is comprised of a support set and a query set.
    Meanwhile, we have a base classes 
    dataset $\mathcal{D}_{base}=\{(x_i,z_i,y_i)|y_i \in C_{base}\}$ which 
    contains abundant examples per base class, and our goal is to learn 
    to classify the examples in the query set $\mathcal{Q}$ with the help of the 
    support set $\mathcal{S}$ and the base classes dataset $\mathcal{D}_{base}$. 
    It should be pointed out that examples from the same class have the same 
    attribute score vector because we only use category-level 
    attribute annotations. 
    
    \subsubsection{Pre-Training Based Meta-Learning Methods}
    Our work follows the line of pre-training based meta-learning methods~\cite{Meta_baseline, FEAT, yang2022sega}.  
    These methods usually train the model in two stages, pre-training and meta-training. And then the learned model is evaluated in the meta-testing stage. Concretely, in the pre-training stage, a feature extractor $f_\theta$ and a classifier are trained on the base classes dataset $\mathcal{D}_{base}$ with standard cross-entropy loss. The pre-trained classifier is removed, and the pre-trained feature extractor $f_{\theta^*}$ is preserved.
    In the meta-training stage, collections of $N$-way $K$-shot tasks are sampled from the base classes
    dataset $\mathcal{D}_{base}$, and each task consists of a support set and a query set. The sampled tasks are used to train a learnable distance metric~\cite{Meta_baseline} or module~\cite{FEAT} with or without fine-tuning the pre-trained feature extractor $f_{\theta^*}$. Finally, in the meta-testing stage, where tasks are sampled from novel classes, the learned model can directly utilize the support set $\mathcal{S}$ to classify the examples on the query set $\mathcal{Q}$ without additional training on the support set $\mathcal{S}$.

    \subsection{Component Prototype Learning}
    
    \begin{figure*}[!ht]
        \centering
        \includegraphics[width=0.7\textwidth]{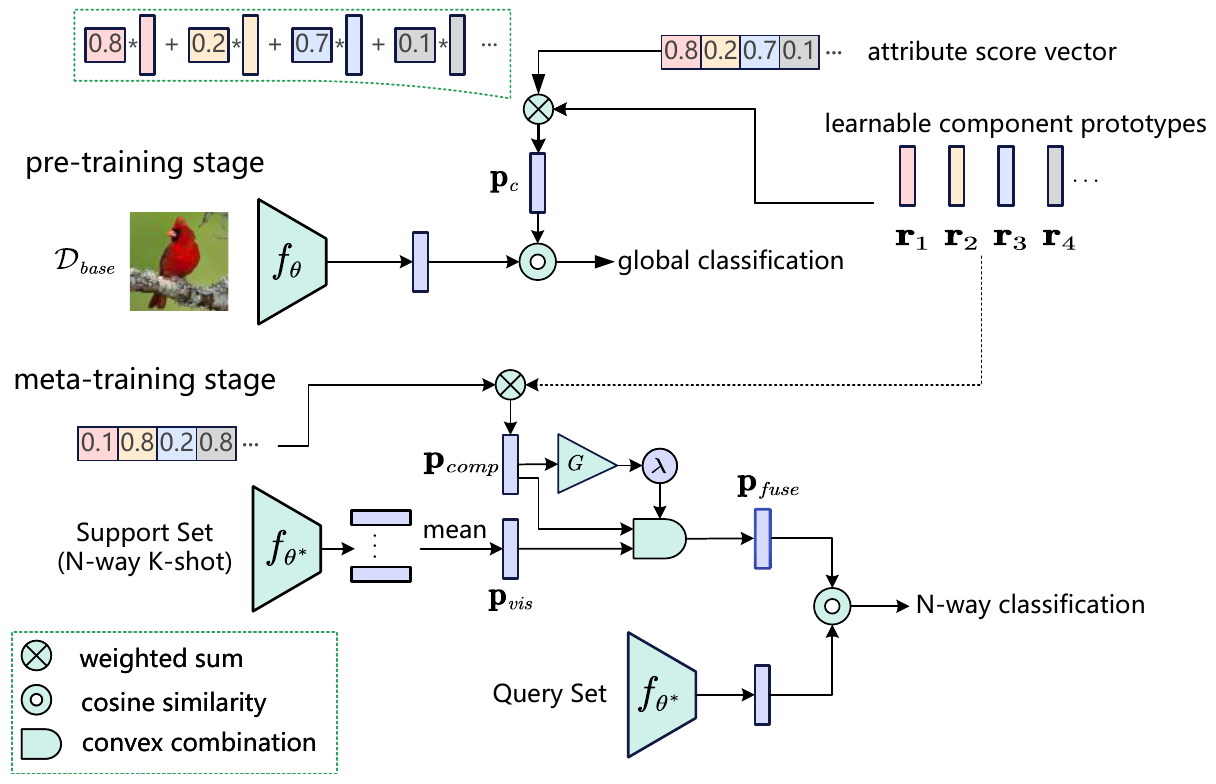}
        \caption{Compositional Prototypical Networks (CPN). In the pre-training stage, CPN trains a feature extractor $f_\theta$ and a set of component prototypes $\{r_{j}\}_{j=1}^M$ for the predefined attribute set $\mathcal{A}=\{a_j\}_{j=1}^M$ by minimizing the global classification loss for examples from the base classes dataset $\mathcal{D}_{base}$. The class prototype $p_c$ is constructed as a weighted sum of 
        the L2-normalized component prototypes. Then in the meta-training stage, the learned component prototypes are reused to construct the compositional prototype $p_{comp}$. Meanwhile, CPN calculates the mean feature for examples of each class in the support set as the visual prototype $p_{vis}$. A learnable weight generator $G$ is used to adaptively fuse the compositional and visual prototypes. The fused prototype $p_{fuse}$ is used to classify examples from the query set with cosine similarity as the distance metric.}
        \label{figure_CPN}
    \end{figure*}
    
    Different from most existing methods~\cite{Meta_baseline, FEAT} which only reuse the feature extractor
    learned in the pre-training stage, we propose \textit{Compositional Prototypical Networks} to learn a prototype for each attribute $a_j$ in the 
    predefined attribute set  $\mathcal{A}=\{a_j\}_{j=1}^M$, which we call a component prototype. These component prototypes can be viewed as a kind of meta-knowledge since they can be shared across different learning tasks. Specifically, in the meta-training and meta-testing stage, they can be reused to construct compositional prototypes for classes from the support set.
    
    As shown in Figure~\ref{figure_CPN}, in the pre-training stage, we define 
    a set of learnable component prototypes $\{\mathbf{r}_{j}\}_{j=1}^M$ for the predefined attribute set $\mathcal{A}=\{a_j\}_{j=1}^M$, where $\mathbf{r}_{j}\in \mathbb{R}^d$ denotes the component prototype for the attribute $a_j$. Then we utilize the learnable component prototypes $\{\mathbf{r}_{j}\}_{j=1}^M$ and category-level
    attribute score vector $z_c$ to calculate
    the class prototype $\mathbf{p}_c$ for each class  $c\in C_{base}$. Concretely, we first perform
    L2-normalization on the learnable component prototypes $\{\mathbf{r}_{j}\}_{j=1}^M$ to 
    obtain the L2-normalized component prototypes $\{\hat{\mathbf{r}}_{j}\}_{j=1}^M$, where
    $\hat{\mathbf{r}}_{j}=\mathbf{r}_{j}/||\mathbf{r}_{j}||_2$. Then we calculate the class prototype $\mathbf{p}_c$ as a weighted sum of 
    the L2-normalized component prototypes $\{\hat{\mathbf{r}}_{j}\}_{j=1}^M$:
    \begin{equation}
      \label{eq_class_proto}
      \mathbf{p}_c=\sum_{j=1}^M{z_{c,j}\hat{\mathbf{r}}_{j}},
    \end{equation}
    where $z_{c,j}$ denotes the score for class $c$ on attribute $a_j$.
    
    Given an image $x_i$ from the base classes dataset $\mathcal{D}_{base}$, we use cosine similarity to calculate
    the probability that $x_i$ belongs to class $c$ as follows:
    \begin{equation}
      \label{eq_probability}
      P(\hat{y}=c\mid x_i)=\frac{\exp \left( \tau_1 \cdot \langle f_{\theta}(x_i),\mathbf{p}_c\rangle \right)}{\sum_{c'}{\exp}\left( \tau_1 \cdot \langle f_{\theta}(x_i),\mathbf{p}_{c'}\rangle \right)},
    \end{equation}
    where $f_{\theta}(x_i)\in \mathbb{R}^d$, and $\tau_1$ is a learnable temperature parameter, and $\langle \cdot,\cdot \rangle$ denotes cosine similarity operator.
    The classification loss for the whole base classes dataset $\mathcal{D}_{base}$ 
    can be defined as follows using cross-entropy loss:
    \begin{equation}
      \label{eq_loss}   
      \mathcal{L} =-\frac{1}{\left| \mathcal{D} _{base} \right|}\sum_{\left( x_i,z_i,y_i \right) \in \mathcal{D} _{base}}{\log P\left( \hat{y}=y_i|x_i \right)}.
    \end{equation}
    The model is trained by minimizing Equation \ref{eq_loss} and 
    then we can obtain learned component prototypes $\{\mathbf{r}_{j}^*\}_{j=1}^M$, which can 
    be regarded as transferable meta-knowledge distilled from the base classes dataset $\mathcal{D}_{base}$.
    
    \subsection{Adaptive Prototype Fusion}
    In the meta-training (episodic training) stage, a collection of $N$-way $K$-shot tasks sampled from the base classes dataset $\mathcal{D}_{base}$ are given to 
    mimic the scenario in the meta-testing stage. In this stage, we focus on learning
    an adaptive weight generator~\cite{MAP, AM3,ma2022adaptive}, which is used to fuse two types of prototypes, namely, a compositional prototype $\mathbf{p}_{comp}$ and a visual prototype $\mathbf{p}_{vis}$. To simplify the notation, we choose one of the $N$ classes from the support set to illustrate the whole fusion process. That is to say, $\mathbf{p}_{comp}$ and $\mathbf{p}_{vis}$ correspond to any class $s$ from the support set.
    
    Similar to Equation \ref{eq_class_proto}, we reuse the learned component prototypes
    to construct the compositional prototype $\mathbf{p}_{comp}$ for class $s$:
    \begin{equation}
      \label{eq_comp_proto}
      \mathbf{p}_{comp}=\sum_{j=1}^M{z_{s,j}\hat{\mathbf{r}}_{j}^*},
    \end{equation}
    where $\hat{\mathbf{r}}_{j}^{*}=\mathbf{r}_{j}^{*}/||\mathbf{r}_{j}^{*}||_2$, and $z_{s,j}$ denotes the score for class $s$ on attribute $a_j$. Following Prototypical Networks~\cite{Prototypical_Net}, we calculate the visual prototype $\mathbf{p}_{vis}$ 
    for class $s$ as the mean vector of the extracted features for examples labeled with class $s$ in the support set:
    \begin{equation}
      \label{eq_vis_proto}
      \mathbf{p}_{vis}=\frac{1}{K}\sum_{i\in \left\{ i|y_i=s \right\}}{f_{\theta^*}}\left( x_i \right), 
    \end{equation}
    where $f_{\theta^*}$ denotes the pre-trained feature extractor, and $K$ is the number of examples belonging to class $s$ since the support set is sampled using 
    $N$-way $K$-shot setting.
    
    Before we fuse the above two prototypes, we need to L2-normalize them. Following 
    ~\cite{MAP,AM3},
    we employ a convex combination to fuse the L2-normalized compositional prototype
    $\mathbf{\hat{\mathbf{p}}}_{comp}$ and the L2-normalized visual prototype $\mathbf{\hat{\mathbf{p}}}_{vis}$, where $\mathbf{\hat{\mathbf{p}}}_{comp}=\mathbf{p}_{comp}/||\mathbf{p}_{comp}||_2$
    and $\mathbf{\hat{\mathbf{p}}}_{vis}=\mathbf{p}_{vis}/||\mathbf{p}_{vis}||_2$.
    We use a learnable weight generator $G$ followed by a sigmoid function to adaptively generate a weight coefficient $\lambda$ and use  the coefficient $\lambda$ to calculate the fused prototype $\mathbf{p}_{fuse}$ as follows:
    \begin{equation}
      \label{eq_weight}
      \lambda =\frac{1}{1+\exp \left( -G\left( \hat{\mathbf{p}}_{comp} \right) \right)},
    \end{equation}
    \begin{equation}
      \label{eq_fuse_proto}
      \mathbf{p}_{fuse}=\lambda \hat{\mathbf{p}}_{comp}+\left( 1-\lambda \right) \hat{\mathbf{p}}_{vis}.
    \end{equation}
    
    The sigmoid function used in Equation \ref{eq_weight} is to make sure the weight coefficient $\lambda$ is between 0 and 1 so that the right side in Equation \ref{eq_fuse_proto} is a convex combination.
    Given an image $q_i$ from the query set, we still use cosine similarity to predict the probability that $q_i$ belongs to class $s$ as follows:
    \begin{equation}
      \label{eq_query_probability}
      P(\hat{y}=s\mid q_i)=\frac{\exp \left( \tau_2 \cdot \langle f_{\theta ^*}(q_i),\mathbf{p}_{fuse}^{s}  \rangle \right)}{\sum_{s'}{\exp}\left( \tau_2 \cdot  \langle f_{\theta ^*}(q_i),\mathbf{p}_{fuse}^{s'}  \rangle \right)},
    \end{equation}
    where $\mathbf{p}_{fuse}^{s'}$ denotes the fused prototype for class $s'$, and $\tau_2$ is another learnable temperature parameter.
    
    In the meta-training stage, the learnable weight generator $G$ and the learnable temperature parameter $\tau_2$ are optimized
    by minimizing the classification loss of examples in the query set. It is worth noting that we \textbf{further} optimize the learned component prototypes in the meta-training stage. This implementation is based on experimental results. 
    Finally, in the meta-testing stage, the learned model calculates the fused prototype for each novel class from the support set $\mathcal{S}$ following Equation \ref{eq_comp_proto}-\ref{eq_fuse_proto}. The calculated prototypes can be directly used to classify examples in the query set $\mathcal{Q}$ according to Equation \ref{eq_query_probability}.

    \section{Experiments}
    \subsection{Experimental Setup}
    \paragraph{Datasets.}
    We conduct the experiments on two datasets with
    human-annotated attribute annotations: Caltech-UCSD-Birds 200-2011 (CUB)~\cite{wah2011caltech}, and SUN Attribute Database (SUN)~\cite{patterson2014sun}.
    CUB is a fine-grained bird species dataset containing 11,788 bird images from 200 species and 312 predefined attributes. There is only one attribute vector for each class in CUB, which we call a category-level attribute score vector.
    SUN is a scene recognition dataset containing 14,340 images from 717 categories and 102 predefined attributes. It should be noted that each image in SUN has an attribute score vector, which we call an image-level attribute score vector. 
    As we have mentioned earlier, we only use category-level attribute score vectors. To this end, we calculate the mean vector of image-level attribute score vectors from the same class as this class's attribute score vector.
    
    \paragraph{Experimental Settings.}
    Our experiments are conducted in 5-way 1-shot and 5-way 5-shot settings. As~\cite{A_closer_look}, we divide CUB into 100 training classes, 50 validation classes, and 50 testing classes.
    As~\cite{AGAM_attention_alignment,MAP}, we divide SUN into 580 training classes, 65 validation classes, and 72 testing classes.
    We sample 15 query examples per class in each task for the meta-training and meta-testing stages. We report the \textit{average accuracy} (\%) and the corresponding 95\% \textit{confidence intervals} over 5000 test episodes to make a fair comparison. Our work follows the inductive setting, where each example in the query set is classified independently.
    
    \paragraph{Implementation Details.}
    Our experiments are conducted using the convolution neural network ResNet12~\cite{Meta_baseline}, a popular feature extractor in recent few-shot learning methods. We also give our results using Conv4~\cite{Matching_Net} on SUN for fair comparisons since we find that almost all previous works~\cite{AGAM_attention_alignment, MAP, ASL, AM3} choose Conv4 as their feature extractor on SUN.
    Moreover, we use a simple fully-connected layer as the learnable weight generator $G$, and the temperature parameter $\tau_1$ and $\tau_2$ are initialized as 10.
    We use the SGD optimizer with a momentum of 0.9 and weight decay of $5 \times 10^{-4}$.
    Following~\cite{yang2022sega}, we adopt the random crop, random horizontal flip and erasing, and color jittering to perform data augmentation.
    Dropblock~\cite{ghiasi2018dropblock} regularization is used to reduce overfitting.
    In the pre-training stage, we train the feature extractor for 30
    epochs. In the meta-training stage, we train our model for 10 epochs. The best model is chosen according to the accuracy on the validation set.
    
    \subsection{Comparison to the State-of-the-Art}
    Table \ref{tab_CUB_SOTA} and \ref{table_SUN_SOTA} show the results of our method and previous state-of-the-art methods on CUB and SUN, respectively. It can be observed that the proposed CPN achieves the best performance among all approaches in both 5-way 1-shot and 5-way 5-shot settings.
    We notice that the performance gains are remarkable in the 5-way 1-shot setting. It indicates that the compositional prototype constructed by the learned component prototypes plays a significant role, especially when the visual prototype is inaccurate due to the limited examples in the 5-way 1-shot setting.
    
    Moreover, we notice that better performance can be achieved by using a stronger feature extractor.
    Concretely, when we replace Conv4 with ResNet12 on SUN, we can obtain 7.62\% and 7.65\% performance gains in 5-way 1-shot
    and 5-way 5-shot settings, respectively. It shows that our method can benefit a lot from a more powerful feature extractor.
    
    We attribute the effectiveness of our method to two aspects. The first is that the learned
    component prototypes have good class transferability. By reusing these transferable component prototypes to construct
    the compositional prototype for a novel class, we can roughly get the class center for this class. The second is that
    the adaptive fusion of the compositional and visual prototypes can further improve the classification accuracy by combining visual and semantic information.
    To illustrate these points, we perform carefully designed ablation experiments in Ablation Study.
    \begin{table*}[!ht]
      \centering
        \scalebox{0.95}{
        \begin{tabular}{lccc}
        \toprule
        \multirow{2}[2]{*}{Method} & \multirow{2}[2]{*}{Backbone} & \multicolumn{2}{c}{Test Accuracy} \\
              &       & 5-way 1-shot & 5-way 5-shot \\
        \midrule
        MatchingNet~\cite{Matching_Net} $^\ddagger$ & ResNet12 & 60.96 $\pm$ 0.35 & 77.31 $\pm$ 0.25 \\
        ProtoNet~\cite{Prototypical_Net} & ResNet12 & 68.8  & 76.4 \\       
        FEAT~\cite{FEAT} & ResNet12 & 68.87 $\pm$ 0.22 & 82.90 $\pm$ 0.15 \\
        MAML~\cite{MAML_Finn} & ResNet18 & 69.96 $\pm$ 1.01 & 82.70 $\pm$ 0.65 \\
        AFHN ~\cite{li2020adversarial}   &ResNet18     & 70.53 $\pm$ 1.01    & 83.95 $\pm$ 0.63 \\
        CPDE~\cite{primitive_discovery} &ResNet18 &80.11 $\pm$ 0.34 &89.28 $\pm$ 0.33 \\
        BlockMix~\cite{tang2020blockmix} &ResNet12 &75.31 $\pm$ 0.79 &88.53 $\pm$ 0.49 \\
        DeepEMD~\cite{DeepEMD} &ResNet12 &75.65 $\pm$ 0.83  &88.69 $\pm$ 0.50 \\
        AGPF~\cite{tang2022learning} &ResNet12 &78.73 $\pm$ 0.84 &89.77 $\pm$ 0.47 \\
        MetaNODE~\cite{zhang2022metanode} & ResNet12 & 80.82 $\pm$ 0.75 & 91.77 $\pm$ 0.49 \\
        Comp.~\cite{Comp_rep} $^\star$ & ResNet10 & 53.6  & 74.6 \\
        Dual TriNet~\cite{chen2019multi} $^\star$ $^\circ$ & ResNet18 & 69.61 $\pm$ 0.46 & 84.10 $\pm$ 0.35 \\
        
        AM3~\cite{AM3} $^\star$ & ResNet12 & 73.6  & 79.9 \\
        Multiple-Semantics~\cite{schwartz2022baby} $^\star$ $^\circ$ $^\bullet$ & DenseNet121 & 76.1  & 82.9 \\
        AGAM~\cite{AGAM_attention_alignment} $^\star$ & ResNet12 & 79.58 $\pm$ 0.25 & 87.17 $\pm$ 0.23 \\
        ASL~\cite{ASL} $^\star$ &ResNet12 &82.12 $\pm$ 0.14 &89.65 $\pm$ 0.11 \\
        SEGA~\cite{yang2022sega} $^\star$ & ResNet12 &
        84.57$\pm$0.22 & 90.85 $\pm$ 0.16 \\
        MAP-Net~\cite{MAP} $^\star$ &ResNet12     & 82.45 $\pm$ 0.23    & 88.30 $\pm$ 0.17 \\
        \midrule
        \textbf{CPN (Ours)} $^\star$ & ResNet12  & \textbf{87.29} $\pm$ \textbf{0.20} & 	\textbf{92.54} $\pm$ \textbf{0.14} \\
        \bottomrule  
        \end{tabular}}
      \caption{Comparison with state-of-the-art methods 
      on CUB. We report the average accuracy (\%) with 95\% confidence intervals over 5000 test episodes.
      $^\star$~denotes that it uses attribute annotations.  
      $^\circ$~denotes that it uses word embeddings. 
      $^\bullet$~denotes that it uses natural language descriptions.
      $^\ddagger$~denotes that the results are reported in ~\cite{AGAM_attention_alignment}.
      Best results are displayed in boldface.}
      \label{tab_CUB_SOTA}%
    \end{table*}%
    
    \begin{table*}[!ht]
        \centering
      \scalebox{0.95}{
      \begin{tabular}{lccc} 
        \toprule
        \multirow{2}{*}{Method} & \multirow{2}{*}{Backbone} & \multicolumn{2}{c}{Test Accuracy} \\
         &  & 5-way 1-shot & 5-way 5-shot \\ 
        \midrule
        RelationNet~\cite{Relation_Network}$^\ddagger$ & Conv4 & 49.58 $\pm$ 0.35 & 76.21 $\pm$ 0.19 \\
        MatchingNet~\cite{Matching_Net}$^\ddagger$ & Conv4 & 55.72 $\pm$ 0.40 & 76.59 $\pm$ 0.21 \\
        ProtoNet~\cite{Prototypical_Net}$^\ddagger$ & Conv4 & 57.76 $\pm$ 0.29 & 79.27 $\pm$ 0.19 \\
        Comp.~\cite{Comp_rep}$^\star$ & ResNet10 & 45.9 & 67.1 \\
        AM3~\cite{AM3}$^\star$ $^\ddagger$ & Conv4 & 62.79 $\pm$ 0.32 & 79.69 $\pm$ 0.23 \\
        AGAM~\cite{AGAM_attention_alignment}$^\star$ & Conv4 & 65.15 $\pm$ 0.31 & 80.08 $\pm$ 0.21 \\
        ASL~\cite{ASL}$^\star$ & Conv4 & 66.17 $\pm$ 0.17 & 80.91 $\pm$ 0.15 \\
        MAP-Net~\cite{MAP} $^\star$ & Conv4 & 67.73 $\pm$ 0.30 & 80.30 $\pm$ 0.21 \\ 
        \midrule
        \textbf{CPN-Conv4 (Ours)} $^\star$ & Conv4 & \textbf{80.45} $\pm$ \textbf{0.22} & \textbf{81.56} $\pm$ \textbf{0.21} \\
        \textbf{CPN (Ours)} $^\star$ & ResNet12 & \textbf{88.07} $\pm$ \textbf{0.17} & \textbf{89.21} $\pm$ \textbf{0.15} \\
        \bottomrule
      \end{tabular}}
      \caption{Comparison with state-of-the-art methods 
      on SUN. We report the average accuracy (\%) with 95\% confidence intervals over 5000 test episodes.
      $^\star$~denotes that it uses attribute annotations.
      $^\ddagger$~denotes that the results are reported in~\cite{AGAM_attention_alignment}.
      Best results are displayed in boldface.}
      \label{table_SUN_SOTA}
    \end{table*}

    \begin{figure}[!ht]
      \centering
      \subfigure[LCP]{
      \label{TSNE_LCP}
      \includegraphics[width=0.45\columnwidth]{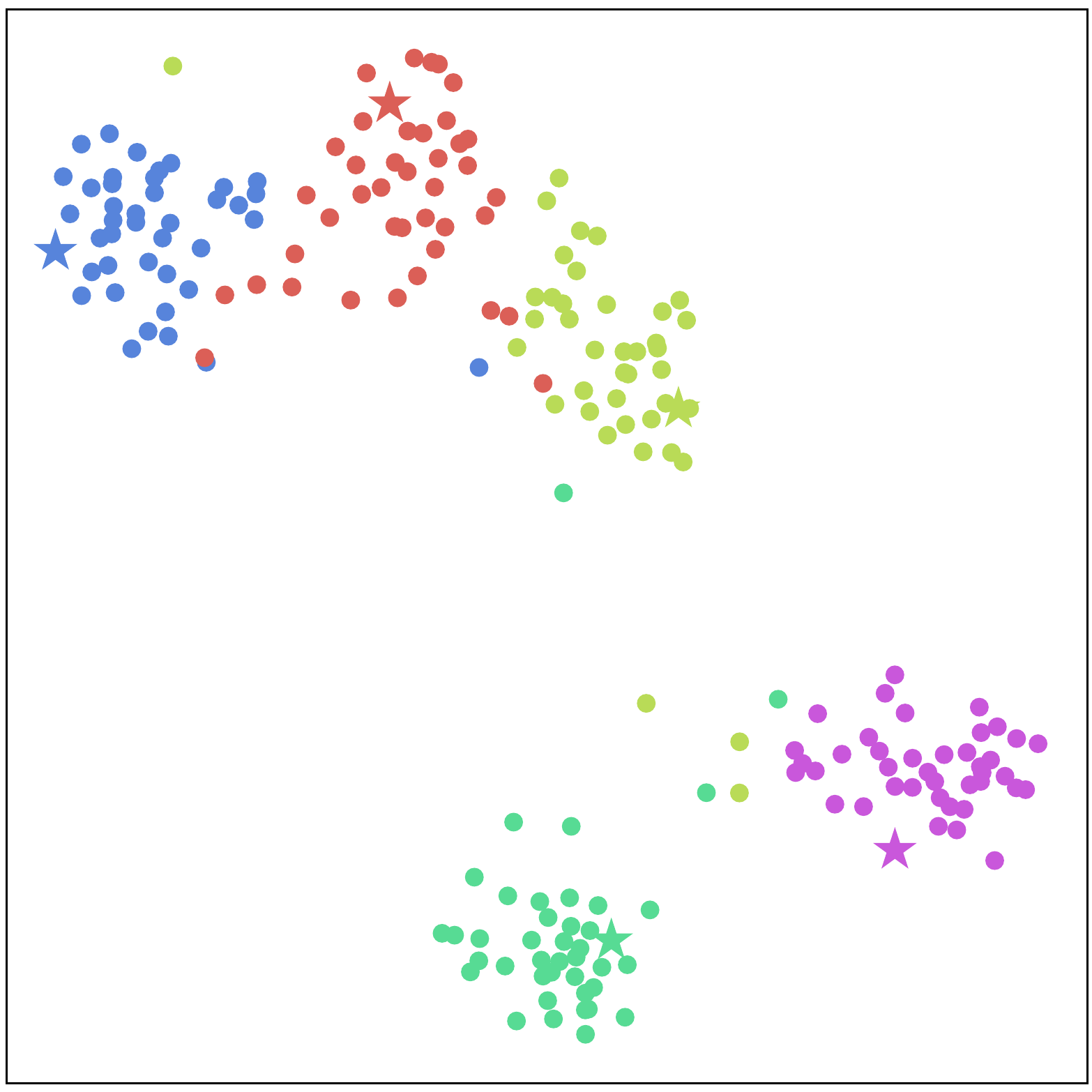}
      }
      \quad
      \subfigure[LCP+]{
      \label{TSNE_LCP_MT}
      \includegraphics[width=0.45\columnwidth]{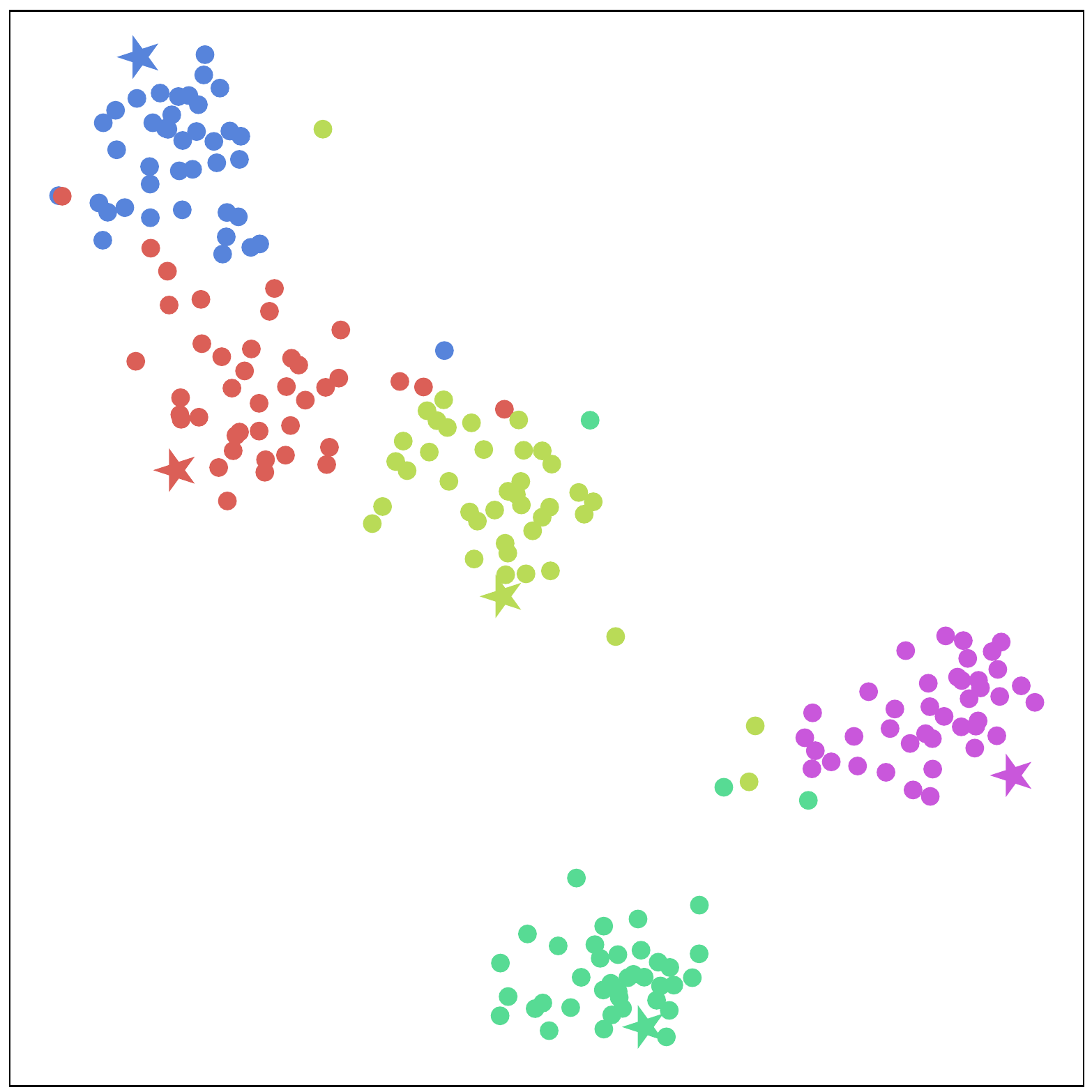}
      }
      \quad
      \subfigure[VP]{
      \label{TSNE_VP}
      \includegraphics[width=0.45\columnwidth]{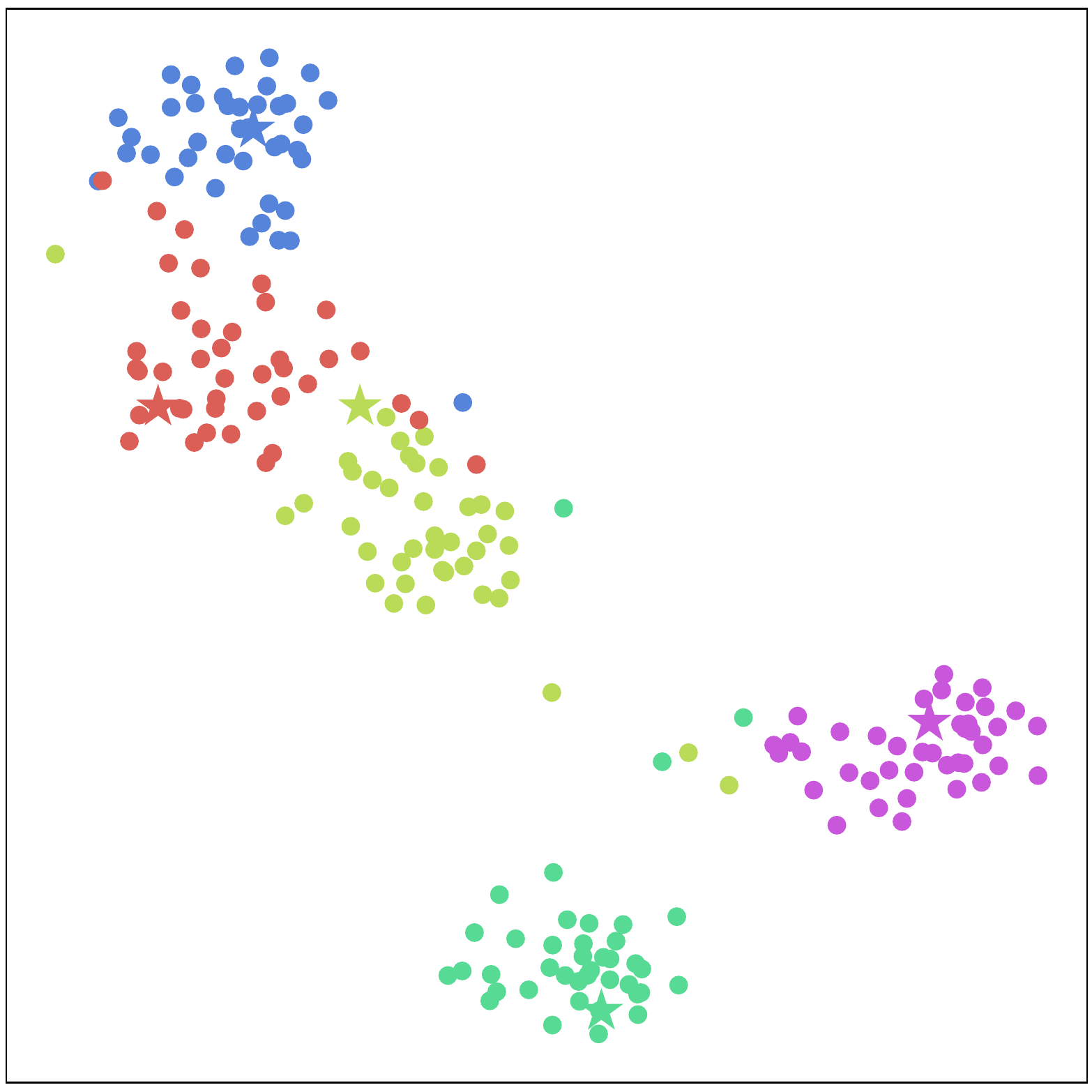}
      }
      \quad
      \subfigure[Ours (VP and LCP+)]{
      \label{TSNE_Ours}
      \includegraphics[width=0.45\columnwidth]{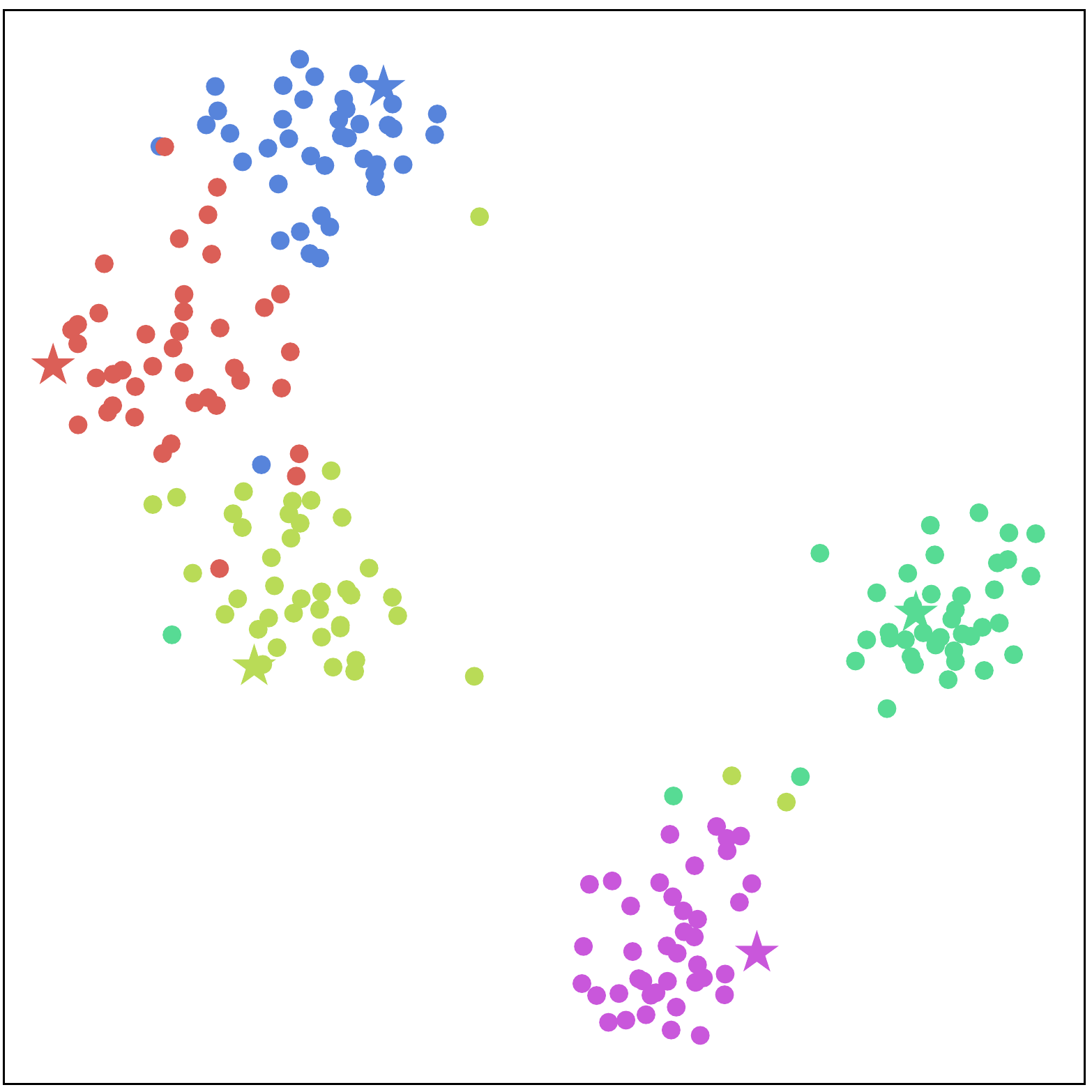}
      }
      \caption{The t-SNE visualization using different prototypes (See Ablation Study for the meaning of VP, LCP and LCP+).
      '$\star$' denote prototypes for different classes, and '$\bullet$' denotes features for the query set. 
      We sample 40 examples from each class to form the query set for a better view.}
      \label{figure_TSNE}
    \end{figure}
    \subsection{Ablation Study}
    
    \begin{table*}[!ht]
        \centering
      \scalebox{0.90}{
      \begin{tabular}{ccccccc} 
        \toprule
        \multirow{2}{*}{RICP} & \multirow{2}{*}{VP} & \multirow{2}{*}{LCP} & \multicolumn{2}{c}{CUB} & \multicolumn{2}{c}{SUN} \\
         &  &  & 5-way 1-shot & 5-way 5-shot & 5-way 1-shot & 5-way 5-shot \\ 
        \midrule
        $\checkmark$ &  &  & 19.43 $\pm$ 0.21 & 17.44 $\pm$ 0.19 & 20.96~$\pm$ 0.20 & 18.06~$\pm$ 0.18 \\
         & $\checkmark$ &  & 79.62 $\pm$ 0.27 & 92.11 $\pm$ 0.14 & 71.21 $\pm$ 0.29 & 86.61 $\pm$ 0.18 \\
         &  & $\checkmark$ & 82.04 $\pm$ 0.24 & 82.10 $\pm$ 0.24 & 85.63 $\pm$ 0.19 & 85.53 $\pm$ 0.19 \\
        $\checkmark$ & $\checkmark$ &  & 79.66 $\pm$ 0.27 & 92.13 $\pm$ 0.14 & 71.21 $\pm$ 0.29 & 86.60 $\pm$ 0.18 \\
         & $\checkmark$ & $\checkmark$ & 84.54 $\pm$ 0.22 & 89.34 $\pm$ 0.17 & 86.54 $\pm$ 0.18 & 88.28 $\pm$ 0.16 \\ 
        \midrule
        \multicolumn{3}{c}{[VP, LCP+]} &85.27 $\pm$ 0.21  &90.49 $\pm$ 0.16 &86.61 $\pm$ 0.18 &88.31 $\pm$ 0.16 \\
        \multicolumn{3}{c}{Ours (VP and LCP+)} & \textbf{87.29} $\pm$ \textbf{0.20} & \textbf{92.54} $\pm$ \textbf{0.14} & \textbf{88.07} $\pm$ \textbf{0.17} & \textbf{89.21} $\pm$ \textbf{0.15} \\
        \bottomrule
      \end{tabular}}
      \caption{ Ablation experiments on using different prototypes.VP, the \textit{V}isual \textit{P}rototype.
      RICP, the compositional prototype constructed by \textit{R}andomly \textit{I}nitialized \textit{C}omponent \textit{P}rototypes. LCP, the compositional prototype constructed by \textit{L}earned \textit{C}omponent \textit{P}rototypes. 
      LCP+ means the learned component prototypes will be further optimized in the meta training stage. [VP, LCP+], naive fusion using the concatenation of VP and LCP+. Ours, adaptive fusion of VP and LCP+.}
      \label{table_proto_ablation}
    \end{table*}
    
    \begin{table*}[!ht]
        \centering
      \scalebox{0.90}{
      \begin{tabular}{ccccc} 
        \toprule
        \multirow{2}{*}{Input} & \multicolumn{2}{c}{CUB} & \multicolumn{2}{c}{SUN} \\
         & 5-way 1-shot & 5-way 5-shot & 5-way 1-shot & 5-way 5-shot \\ 
        \midrule
        $[\hat{\mathbf{p}}_{vis},\hat{\mathbf{p}}_{comp}]$ & 87.01~$\pm$ 0.20  & \textbf{92.77} $\pm$ \textbf{0.13}  & 87.26~$\pm$ 0.17  & \textbf{89.47~$\pm$ 0.15}  \\
        $\hat{\mathbf{p}}_{vis}$ & 86.36~$\pm$ 0.21 & 92.56~$\pm$ 0.14 & 87.48~$\pm$ 0.17 & 89.30~$\pm$ 0.15 \\
        $\hat{\mathbf{p}}_{comp}$ & \textbf{87.29} $\pm$ \textbf{0.20} & 92.54~$\pm$ 0.14 & \textbf{88.07} $\pm$ \textbf{0.11} & 89.21~$\pm$ 0.15 \\
        \bottomrule
      \end{tabular}}
      \caption{  Ablation experiments on the input of the weight generator. $[\hat{p}_{vis},\hat{p}_{comp}]$ denotes that the input is the concatenation of the the L2-normalized visual prototype $\hat{p}_{vis}$ and the L2-normalized compositional prototype $\hat{p}_{comp} $.}
      \label{table_generator_ablation}
    \end{table*}
    
    To verify that CPN does learn some meaningful component prototypes, we design several experiments using different prototypes: (1) RICP, the compositional prototype constructed by \textbf{R}andomly \textbf{I}nitialized \textbf{C}omponent \textbf{P}rototypes. (2) VP, the \textbf{V}isual \textbf{P}rototype. (3) LCP, the compositional prototype constructed by \textbf{L}earned \textbf{C}omponent \textbf{P}rototypes. (4) Adaptive fusion of RICP and VP. (5) Adaptive fusion of LCP and VP. (6) Naive fusion using the concatenation of VP and LCP+. We use a fully connected layer to reduce the dimension of the concatenation so that it has the same dimension as the visual feature. (7) Ours, adaptive fusion of VP and LCP+. LCP+ in (6) and (7) means the learned component prototypes will be further optimized in the meta training stage. The results are shown in Table \ref{table_proto_ablation}.  
    
    (i) According to the first and third row of Table \ref{table_proto_ablation}, 
    we know that the learned component prototypes do have good class transferability since LCP achieves 
    quite good performance. In contrast, RICP seems meaningless since its classification accuracy is almost equivalent to a random classifier (20\%). 
    (ii) From the second and fourth row of Table \ref{table_proto_ablation}, we learn that RICP 
    still does not work even if we adaptively fuse RICP and VP. By contrast, the classification accuracy in the fifth row 
    is generally higher than both the second and third row in Table \ref{table_proto_ablation}. It suggests the adaptive fusion of LCP and VP can achieve a better performance in most cases. 
    (iii) From the fifth and last row of Table \ref{table_proto_ablation}, we know that better performance can be achieved if we further optimize the learned component prototypes in the meta-training stage.
    (iv) Moreover, from the last and penultimate row of Table \ref{table_proto_ablation}, we learn that the adaptive fusion of VP and LCP+ is better than the naive fusion of VP and LCP+.
    
    In Equation \ref{eq_weight}, we use the L2-normalized compositional prototype $\hat{\mathbf{p}}_{comp}$ as the input 
    of the weight generator $G$ following \cite{AM3}, where the semantic label embedding is used to generate a weight coefficient. However, there are other available choices for the input of the weight generator. 
    The results of related ablation experiments are shown in Table \ref{table_generator_ablation}. 
    We can observe that using $\hat{\mathbf{p}}_{comp}$ as the input of $G$ achieves the best performance in the 1-shot setting. By contrast, using the concatenation of $\hat{\mathbf{p}}_{vis}$
    and $\hat{\mathbf{p}}_{comp}$ achieves the best performance in the 5-shot setting. It may be because $\hat{\mathbf{p}}_{vis}$ provides meaningless information in the 1-shot setting due to the lim                                   ited examples, and using $\hat{\mathbf{p}}_{comp}$ only is more suitable. By contrast, in the 5-shot setting, $\hat{\mathbf{p}}_{vis}$
    summarizes more representative class information from several examples and can provide useful information.
    
    \subsection{Visualization Analysis}
    As shown in Figure~\ref{figure_TSNE}, we use t-SNE~\cite{van2008visualizing} to visualize the feature distributions using different prototypes in the 5-way 1-shot setting on CUB. 
    From Figure~\ref{TSNE_LCP}, we can observe that each class's compositional prototype constructed by the learned component prototypes is well distributed among the examples of that class. This shows that the learned component prototypes have good transferability.
    
    From Figure~\ref{TSNE_VP}, we notice that the visual prototype for the yellow-green class is also very close to the red class, which may result in serious misclassification. This suggests that the visual prototype in the 1-shot setting may not be representative enough due to the limited examples. Figure~\ref{TSNE_Ours} shows the feature distribution using the adaptively fused prototype of VP from Figure~\ref{TSNE_VP} and LCP+ from Figure~\ref{TSNE_LCP_MT}. Compared to Figure~\ref{TSNE_VP}, the adaptively fused prototype for the yellow-green class is pulled away from the red class, which benefits the final classification.
    
    \section{Conclusion}
    In this work, we propose a novel Compositional Prototypical Network (CPN) to learn 
    component prototypes for predefined attributes in the pre-training stage. The learned 
    component prototypes can be reused to construct a compositional prototype for each class in the support set. And in the meta-training stage, we further optimize the learned component prototypes and learn an adaptive weight generator to fuse the compositional and visual prototypes. We empirically show that the learned component prototypes have good class transferability. Moreover, we show that the adaptive fusion of the compositional and visual prototypes can further improve classification performance.
    We hope our work can bring more attention and thought to feature reusability and compositional representations in the few-shot learning field.

\section{Acknowledgments}
This work is supported by NSFC projects under Grant 61976201, 
and NSFC Key Projects of International (Regional) Cooperation and Exchanges under Grant 61860206004.
We would like to thank Yuzhong Zhao for inspiring and helpful discussions.

\bibliography{aaai23}

\end{document}